\documentclass[letterpaper, 10 pt, conference]{ieeeconf}  % Comment this line out if you need a4paper

\IEEEoverridecommandlockouts                              % This command is only needed if 
\title{\LARGE \bf
Spatially-anchored Tactile Awareness for Robust Dexterous Manipulation
}

\author{Jialei Huang$^{1,2, *}$, Yang Ye$^{1,3, *}$, Yuanqing Gong$^{1}$,
Xuezhou Zhu$^{1}$, Yang Gao$^{2,4\dagger}$, Kaifeng Zhang$^{1\dagger}$ \\
$^1$Sharpa \quad $^2$Tsinghua University \quad $^3$Wuhan University \quad $^4$Shanghai Qi Zhi Institute \\
$^*$Equal Contribution \quad $^{\dagger}$Equal Advising 
 }
\usepackage{graphicx}
\usepackage{tikz}
\usepackage{multirow}
\usepackage{amsmath} 
\usetikzlibrary{positioning,calc,fit,decorations.pathreplacing,arrows.meta,arrows}
\begin{document}

\maketitle
\thispagestyle{empty}
\pagestyle{empty}

%%%%%%%%%%%%%%%%%%%%%%%%%%%%%%%%%%%%%%%%%%%%%%%%%%%%%%%%%%%%%%%%%%%%%%%%%%%%%%%%

% \begin{abstract}
% Achieving fine-grained dexterous manipulation remains difficult in the sub-millimeter regime: during the terminal alignment phase, occlusions, specular/low-texture surfaces, and concurrent hand--object motion undermine purely visual estimation. While touch provides the necessary local contact evidence, most visuo–tactile policies ingest tactile images as modality embeddings that are \emph{not} tied to the hand's geometry, leaving the spatial semantics of contact underutilized even when joint/state inputs are available. We introduce \textbf{Space-Anchored Tactile Tokens (SATs)}, a hand-centric representation that restores spatial meaning to touch. SATs kinematically align per-fingertip tactile features to the hand's URDF frames and aggregate them as a set representation that is insensitive to finger ordering; the resulting tokens are fused with vision and proprioception by standard policy architectures. SATs let the policy reason about contact \emph{in the hand frame} without explicit object models or pose readouts. We provide a minimal integration recipe (\emph{align} $\rightarrow$ \emph{encode} $\rightarrow$ \emph{fuse} $\rightarrow$ \emph{act}) and evaluate on representative dexterous manipulation tasks. Across benchmarks, SATs improve success rates and reduce completion time relative to strong visuo–tactile baselines. Ablations indicate that hand-frame alignment---rather than additional model capacity---accounts for the majority of the gains.
% \end{abstract}

\begin{abstract}

Dexterous manipulation requires precise geometric reasoning, yet existing visuo-tactile learning methods struggle with sub-millimeter precision tasks that are routine for traditional model-based approaches. We identify a key limitation: while tactile sensors provide rich contact information, current learning frameworks fail to effectively leverage both the perceptual richness of tactile signals and their spatial relationship with hand kinematics. We believe an ideal tactile representation should explicitly ground contact measurements in a stable reference frame while preserving detailed sensory information—enabling policies to not only detect contact occurrence but also precisely infer object geometry in the hand's coordinate system. We introduce SaTA (Spatially-anchored Tactile Awareness for dexterous manipulation), an end-to-end policy framework that explicitly anchors tactile features to the hand's kinematic frame through forward kinematics, enabling accurate geometric reasoning without requiring object models or explicit pose estimation. Our key insight is that spatially-grounded tactile representations allow policies to not only detect contact occurrence but also precisely infer object geometry in the hand's coordinate system. We validate SaTA on challenging dexterous manipulation tasks, including bimanual USB-C mating in free space—a task demanding sub-millimeter alignment precision—as well as light bulb installation requiring precise thread engagement and rotational control, and card sliding that demands delicate force modulation and angular precision. These tasks represent significant challenges for learning-based methods due to their stringent precision requirements. Across multiple benchmarks, SaTA significantly outperforms strong visuo-tactile baselines, improving success rates by up to \textbf{30\%} while reducing task completion times by \textbf{27\%}.

\end{abstract}

\section{INTRODUCTION}
\begin{figure*}[t]
   \centering

   \includegraphics[width=\linewidth]{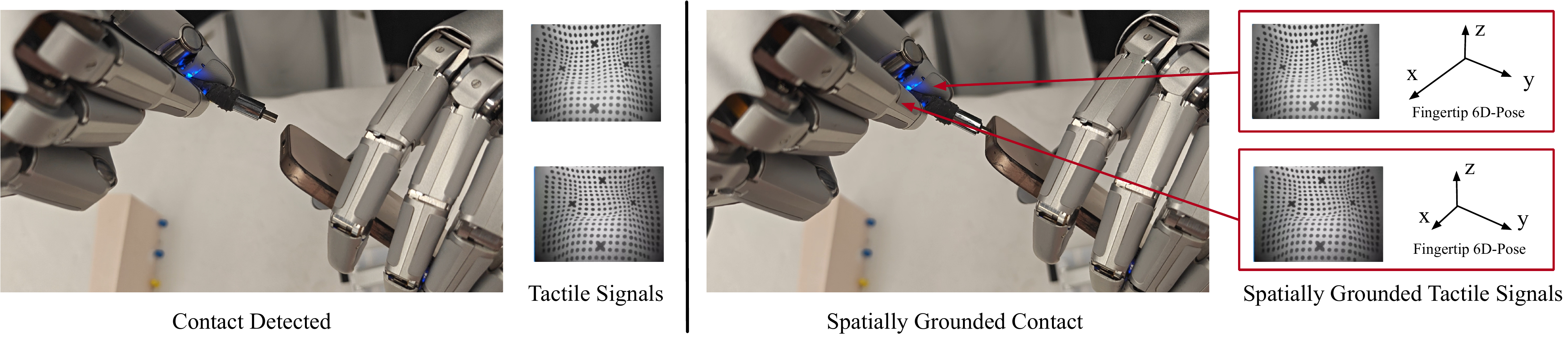}
    \caption{The importance of spatial grounding for tactile-based geometric reasoning. \textbf{Left}: Without spatial grounding, tactile sensors detect contact but cannot infer correct alignment—rich tactile patterns remain isolated measurements. \textbf{Right}: With spatial grounding in the hand's kinematic frame, the same tactile signals enable precise geometric reasoning—each measurement's spatial context guides successful USB-C mating.}
   \label{fig:teaser}
   \vspace{-0.5cm}
\end{figure*}

Dexterous manipulation demands exceptional geometric precision. In multi-finger, multi-contact scenarios, millimeter-level errors can lead to task failure---screws fail to engage threads, USB connectors cannot be inserted, and precision parts cannot be assembled. This challenge becomes particularly severe at the critical moment of contact: extensive finger occlusions block visual observation, specular reflections and textureless surfaces further degrade visual localization accuracy, and the increased complexity from multi-fingered hands amplifies control difficulty. Yet precise geometric information is crucial for successful manipulation at this very moment. When fingers and objects move simultaneously, small perceptual errors accumulate throughout contact-rich interactions, ultimately causing task failure~\cite{c19,c16}.

Vision-based tactile sensors provide critical complementary information for this challenge: they can densely measure contact geometry, deformation distribution, normal/tangential forces, and slip states---precise geometric information that is difficult to reliably obtain through external vision alone~\cite{c1,c2,c3,c4,c6,c7,c8,c9}. However, existing methods have not fully exploited the geometric precision potential of tactile sensing. The core question is: \textbf{how to transform tactile signals into spatially-grounded representations that enable precise geometric reasoning and dexterous manipulation}. Such representations should enable policies to not only know that "contact has occurred," but also understand the local geometric relationships at contact points and reason about necessary adjustments in the hand's coordinate frame. Consider the challenging task of USB-C mating in free space (Figure~\ref{fig:teaser}): when tactile measurements lack proper geometric grounding (left), the policy detects contact but cannot correctly interpret the spatial relationship between tactile patterns—misaligned insertion attempts fail despite rich tactile feedback. In contrast, when the same tactile signals are grounded in the hand's kinematic frame (right), geometric relationships become clear—each tactile measurement carries precise spatial context, enabling accurate alignment and successful mating even under severe visual occlusion~\cite{c16}. This capability to perform geometric reasoning with tactile information is crucial for sub-millimeter precision dexterous manipulation.

Current research follows two main approaches. The first treats geometry reconstruction as an explicit objective, including tactile-based pose estimation~\cite{c10,c11}, tactile SLAM~\cite{c12,c13}, and contact patch estimation~\cite{c14}, but these typically require known object models or complex optimization. The second employs end-to-end learning, feeding tactile signals into policy networks for contact-rich tasks~\cite{c15,c16,c17,c20}. These methods vary in tactile representations—some preserve raw image richness but lack spatial localization~\cite{c15,c17}, while others convert to geometric forms but lose fine details by spatializing tactile features into 3D structures (e.g., point-cloud-style visuotactile representations)~\cite{c18, c35}. An effective tactile representation must reconcile two conflicting requirements: preserving perceptual richness while maintaining spatial grounding for geometric reasoning—this is the core challenge SaTA addresses.

We propose \textbf{SaTA (Spatially-anchored Tactile Awareness for dexterous
manipulation)}, an end-to-end manipulation policy framework based on spatially-anchored tactile sensing. The core of SaTA is anchoring tactile features to the hand's kinematic coordinate system through forward kinematics, while preserving the completeness of geometric information through spatial encoding. This spatially-anchored design enables the policy to accurately infer contact states and object geometry, directly outputting precise manipulation actions without requiring explicit object models or pose estimation.

We validate SaTA on three challenging high-precision dexterous manipulation tasks. \textbf{Bimanual USB-C mating in free space} requires sub-millimeter alignment precision while coordinating two hands without fixed mounting---a significant challenge due to cumulative pose uncertainty. \textbf{Light bulb installation} demands precise thread engagement with controlled force to avoid breakage, combining rotational alignment with compliant insertion. \textbf{Card sliding} necessitates delicate force modulation and angular control to fan cards at specific angles without scattering them, testing fine motor control under minimal visual feedback. These tasks represent substantial challenges for learning-based methods due to their stringent precision requirements and heavy reliance on tactile feedback. Experimental results show that SaTA significantly outperforms strong visuo-tactile baselines across these benchmarks, improving both success rates and reducing completion times.

\textbf{Our main contributions include:}

\begin{enumerate}

\item \textbf{Spatially-anchored tactile representation}: We propose the first representation method that explicitly aligns tactile features with the kinematic frame, simultaneously maintaining perceptual richness and spatial precision;

\item \textbf{End-to-end manipulation policy framework SaTA}: We achieve direct mapping from spatially-anchored tactile perception to precise manipulation actions without intermediate geometric reconstruction steps;

\item \textbf{Challenging task validation}: We demonstrate successful application of learning-based methods on high-precision dexterous manipulation tasks including bimanual USB-C mating, light bulb installation, and card sliding, with ablation studies confirming that spatial anchoring is the key factor for performance improvement.

\end{enumerate}
\section{Method}
\label{sec:method}
\subsection{Overview}
\label{sec:overview}
We consider dexterous manipulation tasks with multi-finger hands, where inputs include tactile images from multiple fingertips, external RGB observations, and robot joint angles, with outputs being target joint angles. The core insight of SaTA is that precise geometric reasoning requires precise tactile measurements to be grounded in a stable reference frame—a capability that existing end-to-end methods lack.

\begin{figure*}[t]
\centering
\includegraphics[width=\linewidth]{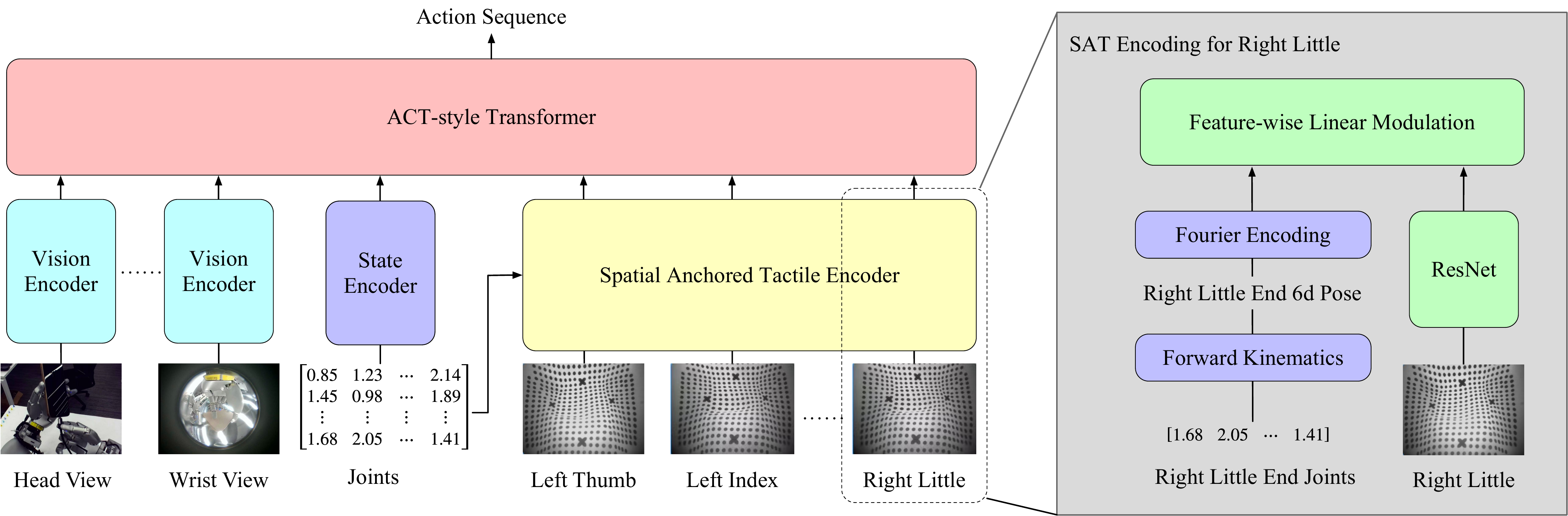}
\caption{\textbf{SaTA Framework Overview.} Our method grounds raw tactile images in the hand's kinematic frame for dexterous manipulation. Multi-modal inputs (head/wrist RGB views, tactile images from fingertips, joint states) are processed through specialized encoders. The core innovation is the Spatial Anchored Tactile Encoder (yellow), which grounds each tactile measurement in the hand's kinematic frame. The detailed SAT Encoding process (gray box) shows how we compute 6D poses via forward kinematics, apply Fourier encoding for multi-scale spatial representation, and use FiLM to integrate spatial context with tactile features from ResNet. All encoded features are processed through an ACT-style Transformer to generate action sequences for precise manipulation.}
\vspace{-5mm}
\label{fig:method_overview}
\end{figure*}

As shown in Figure~\ref{fig:method_overview}, SaTA transforms raw tactile signals into spatially-anchored representations through a unified architecture. The framework processes multi-modal inputs including RGB images from head and wrist cameras, tactile images from fingertip sensors (e.g., Left Thumb, Left Index, Right Little), and robot joint states. The key innovation lies in the Spatial Anchored Tactile Encoder module, which grounds each tactile measurement in the hand's kinematic frame. For each tactile sensor, we: (1) compute its 6D pose via forward kinematics based on current joint angles; (2) encode this spatial information using Fourier features to capture geometric variations at multiple scales~\cite{c39}; (3) integrate spatial context with tactile features through Feature-wise Linear Modulation (FiLM)~\cite{c38}, preserving the rich contact information while adding precise spatial grounding. These spatially-anchored tactile tokens, along with visual encodings and state information, are then processed through an ACT-style Transformer~\cite{c37} to generate action sequences. This design enables geometric reasoning between contacts. It achieves the sub-millimeter precision required for dexterous tasks.

\subsection{Spatially-Anchored Tactile Representation}
\label{sec:spatial_anchoring}
Precise geometric reasoning requires three key elements: a stable reference frame, fine-grained spatial encoding, and information-preserving feature fusion. SaTA systematically realizes these elements through the following design.

\textbf{Choice of Spatial Reference Frame.}
We anchor all tactile measurements to the hand's URDF coordinate system (i.e., the wrist reference frame), rather than the world or camera coordinate systems. This choice is based on a key observation: the success of dexterous manipulation depends on relative geometric relationships \emph{within the hand}, not global positions. For example, when performing USB insertion tasks, regardless of how the wrist moves or the base is positioned, the geometric constraints required for insertion---contact points, alignment angles, insertion directions---remain invariant in the hand coordinate system. This invariance brings two advantages: first, it simplifies policy learning as the policy need not handle irrelevant global motions; second, it provides a natural reference frame for geometric reasoning, allowing manipulation skills learned in one scene to directly transfer to new arm configurations.

\textbf{Fine-grained Spatial Encoding.} 
Sub-millimeter manipulation precision requires the policy to be highly sensitive to spatial variations, particularly in perceiving object pose. For instance, when a phone is tilted at 45 degrees in the air, the USB plug must approach at a matching angle for successful insertion---this requires not only precise position alignment but also accurate perception of relative orientation. To capture both position and orientation changes, we employ Fourier positional encoding~\cite{c39} for the complete 6D pose (3D position + 3D rotation). The key advantage of Fourier features is that their frequency components naturally encode spatial variations at different scales: low-frequency components capture coarse pose relationships (e.g., ``roughly aligned''), while high-frequency components encode fine adjustment needs (e.g., ``rotate 2 degrees'' or ``translate 1mm''). This multi-scale representation enables the network to perform both coarse positioning and fine adjustment simultaneously.

\textbf{Spatially-Aware Feature Fusion.}
Tactile images contain rich local geometric information---texture of contact surfaces, edge orientations, pressure distributions---that is crucial for understanding contact states and cannot be lost during spatial encoding. Therefore, we adopt the FiLM conditioning mechanism~\cite{c38}, which allows spatial information to modulate tactile feature processing rather than simple feature concatenation. The core insight is that the same tactile pattern requires different actions depending on its spatial context. For example, an edge detected at the thumb indicates grip adjustment, while the same edge at the index finger suggests rotation correction. FiLM enables context-dependent interpretation of tactile signals. Each tactile measurement becomes a spatially-anchored token that preserves both sensory information and spatial semantics.
\subsection{Policy Architecture and Training}
\label{sec:policy}
\textbf{Policy Architecture.}
SaTA builds upon the ACT (Action Chunking with Transformers) framework~\cite{c37}, extending its multi-modal inputs to include spatially-anchored tactile information. Specifically, the policy receives three types of inputs at each timestep: (1) robot state information, including current and historical joint angles and end-effector poses; (2) visual information, including RGB-D image features from multiple viewpoints; (3) spatially-anchored tokens from all tactile sensors (e.g., for a 10-fingered tactile system, 10 tactile tokens are input at each timestep regardless of whether contact occurs). We employ a cVAE (Conditional Variational Autoencoder)~\cite{c40} as the state encoder to handle the multi-modality of these inputs and capture the distribution of demonstration data. These multi-modal inputs are processed through a Transformer architecture, where positional encoding ensures the association between tactile tokens and their corresponding fingers. The policy outputs a sequence of actions for the next 100 steps (action chunk), enabling smooth and anticipatory control.

\textbf{Training Strategy.}
SaTA is trained through imitation learning using expert demonstrations collected via human teleoperation. Following the ACT training paradigm, we train the policy by reconstructing action sequences from the cVAE latent representations. During training, tactile features are first anchored to the hand's reference frame, then combined with visual features to learn the mapping from observations to actions. The spatial anchoring provides an explicit geometric inductive bias for the policy network.

It is worth noting that while SaTA is implemented based on the ACT framework in this work, the core idea of spatial anchoring is general and can be easily transferred to other policy architectures. The key innovation---anchoring tactile measurements to the kinematic frame while preserving their spatial semantics---is independent of specific policy network designs, providing a general design principle for tactile-driven dexterous manipulation.
\begin{figure*}[t]
\centering
\includegraphics[width=\linewidth]{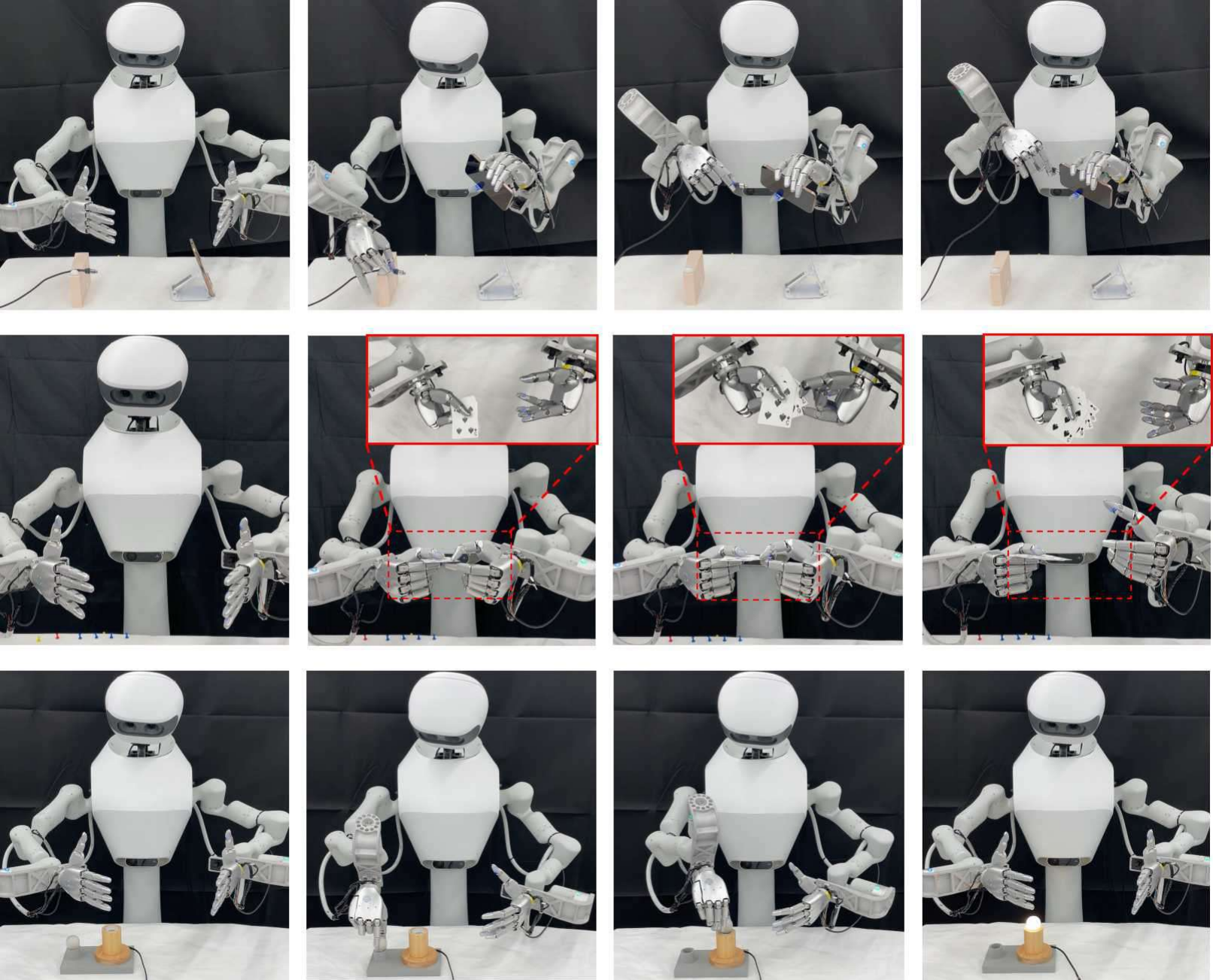}
\caption{\textbf{Task overview and challenges.} Each row represents one task: (top) USB-C mating, (middle) card sliding, (bottom) bulb installation. From left to right, columns show: initial state, approach phase, critical contact moment (red boxes), and successful completion. Red boxes highlight moments where visual information is severely degraded, requiring precise tactile-based geometric reasoning.}
\vspace{-0.3cm}
\label{fig:tasks}
\end{figure*}

\section{Experiments}
\label{sec:experiments}

We design experiments to answer the following key questions: (1) Can SaTA significantly outperform existing methods on dexterous manipulation tasks requiring precise geometric reasoning? (2) How much does each components of spatial anchoring mechanism contribute to performance improvements? (3) What failure modes arise without spatial anchoring, and how does SaTA address them?

\subsection{Experimental Setup}
\label{sec:exp_setup}

\textbf{Task Design.}
We select three manipulation tasks that are extremely sensitive to pose accuracy, representing the capability boundaries of current learning-based manipulation methods (see Figure~\ref{fig:tasks}):

\textit{USB-C Mating:} This bimanual coordination task requires retrieving a USB-C cable and phone from stands, then completing the mating in free space. Each retrieval results in different grasp poses, adding complexity. The critical challenge involves using coordinated rubbing motions between the thumb and index finger to adjust the USB-C plug orientation for alignment. With sub-millimeter positional tolerance and angular tolerance within a few degrees, the plug completely occludes the port during approach, necessitating purely tactile-based geometric reasoning.

\textit{Card Sliding:} This task requires precise control of finger movements to fan out playing cards at specific angles. The key challenge lies in maintaining finger sliding motion along the card surface direction rather than perpendicular to it—perpendicular forces would bend the cards, making them difficult to separate. The task demands sub-millimeter precision and angular tolerance within a few degrees.

\textit{Bulb Installation:} This task involves picking up a light bulb and installing it into a socket. The primary challenge is achieving perpendicular alignment between the bulb and socket threads—even small angular misalignments cause the bulb to tilt against the socket base, preventing successful threading. This task critically depends on accurate pose estimation of the bulb relative to the socket, requiring sub-millimeter precision and angular tolerance within a few degrees while managing appropriate insertion forces.

These tasks share a common characteristic: at the critical moment of contact, visual information is severely degraded or completely unavailable, and small pose errors directly lead to task failure. While traditional engineering solutions rely on high-precision force control or compliance mechanisms, we demonstrate that learning-based methods can achieve comparable precision through spatially-anchored tactile perception.

\textbf{Hardware Platform.}
Experiments are conducted on a dual-arm robotic system equipped with two RealMan 7-DoF manipulators. Each arm is equipped with a Sharpa Wave dexterous hand with 22 degrees of freedom. Each fingertip integrates a vision-based tactile sensor providing 320$\times$240 resolution tactile images at 30Hz. External perception consists of a head-mounted stereo camera and two wrist-mounted fisheye cameras. The system control frequency is 30Hz, synchronized with tactile sampling.

\textbf{Data Collection and Training.}
We collect 200 expert demonstration trajectories for each task through human teleoperation. During teleoperation, operators observe tactile images through a visual interface, with tactile events provided as vibration feedback through the controller. All methods use identical training data, trained for 20 epochs with batch size 64, using the Adam optimizer with learning rate 1e-4. Each experiment is repeated with 3 random seeds to evaluate variance.

\textbf{Evaluation Metrics.}
We evaluate performance using three metrics: (1) \textit{Success Rate (SR)}: percentage of trials where the task is completed successfully within the time limit; (2) \textit{First-Contact Success Rate (FC)}: percentage of trials where correct alignment is achieved on the first contact attempt, measuring the precision of initial pose estimation; (3) \textit{Completion Time}: average time taken to complete successful trials, measured in seconds. The FC metric is particularly important for assessing geometric reasoning capability, as it indicates whether the policy can achieve correct spatial alignment without trial-and-error adjustments.

\subsection{Baseline Methods}
\label{sec:baselines}

We compare SaTA with the following baseline methods:

\textbf{Vision-Only:} A pure vision policy based on the ACT framework~\cite{c37}, using only RGB-D images and joint states as input, without tactile information. This baseline represents a strong benchmark for current visual manipulation methods.

\textbf{Tactile-Flat:} Each of the ten tactile sensors is encoded through a ResNet into a single token, sharing the same ResNet architecture with our method. These ten tokens are concatenated with other modalities without spatial anchoring—essentially our method without the spatial anchoring component.

\textbf{Tactile-Global:} Building upon Tactile-Flat, this baseline incorporates fingertip poses into the proprioceptive data stream rather than anchoring them to individual tactile measurements. While this method has access to the same information as ours, it lacks the explicit spatial grounding of tactile features.

All baselines use identical Transformer architectures and training strategies, differing only in input representation to ensure fair comparison.

\subsection{Experimental Results}
\label{sec:results}
\textbf{Main Results.}
Table~\ref{table:main_results} shows the performance of each method across the three tasks. SaTA significantly outperforms all baseline methods on all tasks, achieving an average success rate of 76.7\% compared to 46.7\% for the strongest baseline. Notably, on the most challenging USB-C mating task, almost all baseline methods fail (0\% success rate), while SaTA achieves 35\% success rate. This gap demonstrates the critical role of spatially-anchored tactile perception for precision manipulation.

% \begin{table}[h]
% \caption{Main Experimental Results Across Three Tasks}
% \label{table:main_results}
% \begin{center}
% \begin{tabular}{|l|c|c|c|c|c|}
% \hline
% \multirow{2}{*}{Task} & \multirow{2}{*}{Metric} & Vision & Tactile & Tactile & \textbf{SaTA} \\
%  & & Only & Flat & Global & \textbf{(Ours)} \\
% \hline
% \multirow{3}{*}{Card Sliding} 
%  & SR (\%) & 50 & 60 & 65 & \textbf{95} \\
%  & FC (\%) & 10 & 45 & 30 & \textbf{55} \\
%  & Time (s) & 24.9 & 16.7 & 18.5 & \textbf{12.7} \\
% \hline
% \multirow{3}{*}{USB-C Mating} 
%  & SR (\%) & 0 & 0 & 10 & \textbf{35} \\
%  & FC (\%) & 0 & 0 & 0 & \textbf{30} \\
%  & Time (s) & - & - & 82.6 & \textbf{58.4} \\
% \hline
% \multirow{3}{*}{Bulb Installation} 
%  & SR (\%) & 45 & 70 & 65 & \textbf{100} \\
%  & FC (\%) & 35 & 40 & 45 & \textbf{60} \\
%  & Time (s) & 272.1 & 195.8 & 160.5 & \textbf{132.7} \\
% \hline
% \multirow{3}{*}{Average} 
%  & SR (\%) & 31.7 & 45.0 & 50.0 & \textbf{76.7} \\
%  & FC (\%) & 15.0 & 21.7 & 25.0 & \textbf{48.3} \\
%  & Time (s) & 148.5 & 109.3 & 89.5 & \textbf{67.9} \\
% \hline
% \end{tabular}
% \end{center}
% \vspace{-2mm}
% \end{table}

\begin{table}[h]
\caption{Main Experimental Results Across Three Tasks}
\label{table:main_results}
\begin{center}
\begin{tabular}{|l|c|c|c|c|c|}
\hline
\multirow{2}{*}{Task} & \multirow{2}{*}{Metric} & Vision & Tactile & Tactile & \textbf{SaTA} \\
 & & Only & Flat & Global & \textbf{(Ours)} \\
\hline
\multirow{3}{*}{Card Sliding} 
 & SR (\%) & 50 & 60 & 65 & \textbf{95} \\
 & FC (\%) & 10 & 45 & 30 & \textbf{55} \\
 & Time (s) & 24.9 & 16.7 & 18.5 & \textbf{12.7} \\
\hline
\multirow{3}{*}{USB-C Mating} 
 & SR (\%) & 0 & 0 & 10 & \textbf{35} \\
 & FC (\%) & 0 & 0 & 0 & \textbf{30} \\
 & Time (s) & - & - & 82.6 & \textbf{58.4} \\
\hline
\multirow{3}{*}{Bulb Installation} 
 & SR (\%) & 45 & 70 & 65 & \textbf{100} \\
 & FC (\%) & 35 & 40 & 45 & \textbf{60} \\
 & Time (s) & 272.1 & 195.8 & 160.5 & \textbf{132.7} \\
\hline
\multirow{3}{*}{Average} 
 & SR (\%) & 31.7 & 43.3 & 46.7 & \textbf{76.7} \\
 & FC (\%) & 15.0 & 28.3 & 25.0 & \textbf{48.3} \\
 & Time (s) & - & - & 87.2 & \textbf{67.9} \\
\hline
\end{tabular}
\end{center}
\vspace{-2mm}
\end{table}
The first-contact success rate---measuring the proportion of trials where correct alignment is achieved on first contact---is a key indicator of pose precision. SaTA's advantage is even more pronounced on this metric (48.3\% vs 25.0\% for the best baseline), indicating that spatial anchoring indeed provides precise geometric reasoning capabilities. Regarding completion time, despite requiring additional computation for processing tactile information, SaTA reduces average completion time by 28\% due to fewer trial-and-error attempts.

\textbf{Ablation Study.}
Table~\ref{table:ablation} presents ablation study results on the card sliding task to analyze key design choices. First, replacing FiLM with simple concatenation (w/o FiLM) reduces success rate by 25\%, demonstrating the importance of spatial modulation for preserving tactile information integrity. Second, removing Fourier encoding (w/o Fourier) causes a 25\% performance drop, confirming that multi-scale spatial encoding is crucial for fine-grained manipulation. Finally, anchoring to the world frame instead of the hand frame (World Frame) results in a 35\% decrease in success rate, validating our choice of the hand's URDF coordinate system for maintaining invariance to arm configurations. Note that the Tactile-Global baseline in Table~\ref{table:main_results} effectively represents our method without spatial anchoring.

\begin{table}[t]
\caption{Ablation Study on Card Sliding Task}
\label{table:ablation}
\begin{center}
\begin{tabular}{|l|c|c|c|}
\hline
Configuration & SR (\%) & FC (\%) & Time (s) \\
\hline
SaTA (Full) & \textbf{95} & \textbf{55} & \textbf{12.7} \\
\hline
w/o FiLM & 70 & 30 & 15.9 \\
\hline
w/o Fourier & 70 & 35 & 22.7 \\
\hline
World Frame & 60 & 25 & 27.4 \\
\hline
\end{tabular}
\end{center}
\end{table}

\begin{figure}[t]
\centering
\includegraphics[width=0.8\columnwidth]{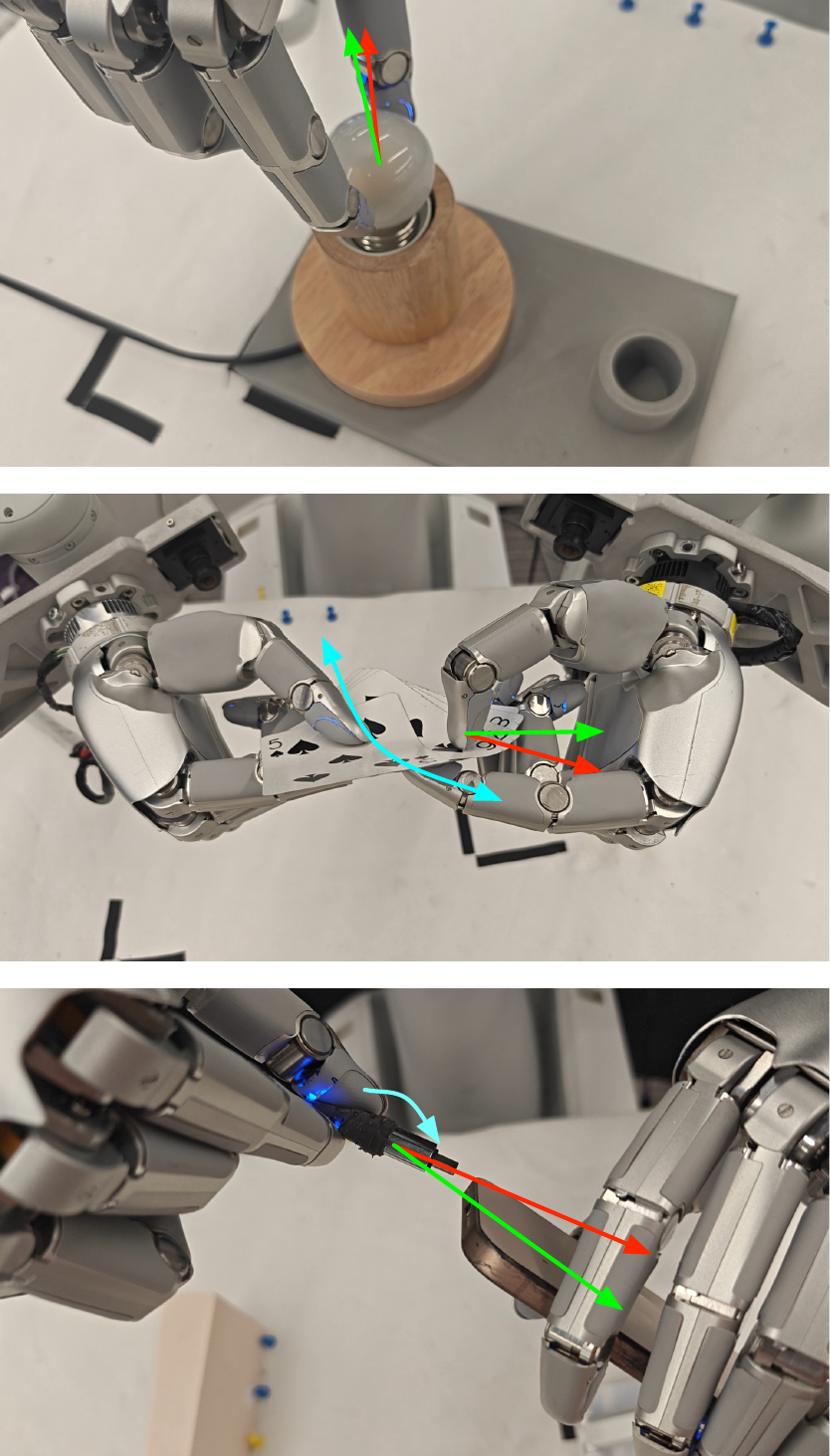}
\caption{Failure mode analysis of Tactile-Flat baseline. Red arrows indicate incorrect directions/orientations executed by the policy, while green arrows show the required correct directions. (Top) Bulb installation: misaligned insertion angle prevents thread engagement. (Middle) Card sliding: perpendicular thumb motion (red) instead of along-surface motion (green) causes card bending (blue curve). (Bottom) USB-C mating: incorrect plug orientation (red) and absence of thumb-index coordinated adjustment motion (blue arrows indicate the required but unlearned motion).}
\vspace{0cm}
\label{fig:failure}
\end{figure}

\textbf{Failure Mode Analysis.}
Figure~\ref{fig:failure} visualizes the geometric precision gap between methods with and without spatial anchoring. The figure shows representative failure cases from the Tactile-Flat baseline, where red arrows indicate incorrect motions/orientations executed by the policy, green arrows show the correct required directions, and blue annotations highlight critical missing behaviors or unintended deformations.

For \textit{bulb installation} (top), the baseline fails to maintain perpendicular alignment during insertion. The red arrow shows the tilted insertion angle learned by the policy, while the green arrow indicates the required vertical direction. This misalignment prevents the bulb from engaging with the socket threads, causing it to jam against the socket rim. Without spatial anchoring, the policy cannot interpret tactile feedback to correct this angular error.

In \textit{card sliding} (middle), the baseline applies force perpendicular to the card surface (red arrow) rather than along it (green arrow), causing the cards to bend (blue curve) instead of sliding apart. This fundamental misunderstanding of the required motion direction stems from the inability to map tactile sensations to appropriate spatial adjustments in the hand's coordinate frame.

The \textit{USB-C mating} task (bottom) reveals the most complex failure mode. The baseline cannot learn the precise thumb-index coordinated rubbing motion (blue arrows) required to adjust the plug orientation from its initial misaligned state (red arrow) to the correct alignment (green arrow). This bimanual coordination requires understanding how tactile patterns on both fingers relate to the plug's orientation—a capability that emerges only when tactile features are spatially grounded.

These failures highlight a critical insight: while tactile sensing without spatial anchoring can detect contact occurrence, it cannot provide the geometric understanding necessary for precise manipulation. The consistent pattern across all three tasks—incorrect force directions, missing corrective behaviors, and inability to maintain precise orientations—demonstrates that spatial anchoring is not merely beneficial but essential for learning dexterous manipulation from demonstration.

In contrast, SaTA's spatially-anchored representation enables the policy to correctly interpret the geometric meaning of tactile patterns, learning both the primary manipulation actions and the subtle corrective behaviors required for task success. The 35\% improvement in success rate over Tactile-Flat directly reflects this enhanced geometric reasoning capability.

\subsection{Discussion}
\label{sec:discussion}

\textbf{Three Levels of Tactile Sensing in Dexterous Manipulation.}
Through this study, we identify three progressive roles that tactile sensing plays in dexterous manipulation:

(1) \textit{Gating signals}: At the most basic level, tactile sensing detects force discontinuities to trigger policy phase transitions. For example, a force spike during USB insertion marks the transition from ``alignment'' to ``insertion'' phase. These signals are extremely simple (no more than 3 bits of information) but crucial for task structure.

(2) \textit{Geometric reasoning}: The focus of this work. Tactile sensing provides high-precision local geometric information, complementing occluded or degraded visual information. Through spatial anchoring, this local information is organized into spatial representations suitable for reasoning, achieving millimeter-level precision control.

(3) \textit{Force-dominant control}: The highest level, where policies are entirely based on force/tactile feedback, with vision providing only coarse guidance. Tasks like pen spinning require continuous force modulation. However, current data collection methods limit the realization of this level.

\textbf{Limitations of Current Approach.}
Our experiments reveal fundamental limitations of the current paradigm: teleoperation data collection cannot provide real tactile feedback. Operators only perceive contact events through vibration, not actual force distributions and textures, resulting in demonstrations that remain vision-dominant with tactile sensing playing only an auxiliary role. This explains why even SaTA has room for improvement on tasks requiring fine force control (e.g., the final tightening phase of bulb installation).

Achieving truly force-dominant policies requires next-generation data collection techniques---such as haptic gloves providing real force feedback, or real-world reinforcement learning allowing robots to autonomously explore force control strategies. These directions represent the future of tactile-driven dexterous manipulation.
\section{Related Work}
\label{sec:related_work}

\textbf{Tactile Sensing for Contact-rich Manipulation.}
Vision-based tactile sensors~\cite{c1,c2,c3,c4,c5,c6} provide rich geometric and force information crucial for manipulation. Tactile-based control methods demonstrate effectiveness across diverse tasks: servoing~\cite{c21}, deformable object manipulation~\cite{c22}, dense packing~\cite{c23}, and assembly under occlusion~\cite{c24,c25}. Recent tactile representation learning~\cite{c28,c29,c36} and vision-touch-language models~\cite{c46,c47,c48,c49,c50} combine multiple modalities for generalizable manipulation, but typically process tactile signals as abstract features without explicit spatial grounding. Methods attempting spatial alignment convert tactile data into simplified geometric forms: Robot Synesthesia~\cite{c18} and 3D-ViTac~\cite{c35} use 3D point clouds, losing contact textures and normal distributions; 
% ViTacFormer~\cite{c36} innovates in cross-modal fusion but lacks explicit spatial anchoring; 
Active Extrinsic Contact Sensing~\cite{c24} compresses to contact lines, over-abstracting geometric information. Unlike these approaches, SaTA preserves complete tactile image features while explicitly anchoring them to the hand's kinematic frame through FiLM modulation, enabling context-dependent interpretation where the same tactile pattern triggers different actions based on spatial position—achieving the precision-performance balance needed for sub-millimeter manipulation.

\textbf{Tactile Sensing for Geometric Reasoning.}
Another research line emphasizes anchoring tactile measurements to specific reference frames for geometric estimation. Explicit methods achieve millimeter-level localization through registration: Bauza et al.'s series from tactile mapping and localization~\cite{c30} to Tac2Pose~\cite{c10} estimate object poses by registering high-resolution tactile images with rendered models. Multi-sensor fusion methods like ViHOPE~\cite{c31} and ViTaSCOPE~\cite{c32} jointly optimize visual point clouds and tactile normals for robust in-hand tracking. Neural implicit representations offer new modeling paradigms: Neural Contact Fields~\cite{c33} track extrinsic contact trajectories, NeuralFeels~\cite{c19} achieves sub-millimeter reconstruction accuracy through visuo-tactile neural fields, and Touch-GS~\cite{c34} uses tactile supervision to improve contact consistency in 3D reconstruction. These methods validate the value of spatial anchoring for geometric precision but focus primarily on perception and reconstruction rather than direct manipulation control. Their reliance on explicit object models or offline optimization procedures limits scalability to diverse manipulation scenarios.

\section{Conclusion}
\label{sec:conclusion}

We presented SaTA, a framework that enables learning-based policies to achieve millimeter-level precision in dexterous manipulation through spatially-anchored tactile representations. Our key insight is that precise geometric reasoning requires tactile measurements to be explicitly anchored to the hand's kinematic frame, preserving both local geometric details and global spatial context. Through extensive experiments on challenging tasks including bimanual USB-C mating, we demonstrated that SaTA significantly outperforms existing visuo-tactile methods, achieving success rates previously unattainable for learning-based approaches.

Our work reveals three progressive levels of tactile sensing in manipulation: gating signals for phase transitions, geometric reasoning for precise alignment, and force-dominant control for compliant interaction. While SaTA successfully addresses the second level through spatial anchoring, current teleoperation-based data collection limits progress toward truly force-dominant policies. Future work should explore haptic interfaces that provide realistic force feedback during demonstration, or leverage real-world reinforcement learning to discover force control strategies autonomously.

The success of spatial anchoring suggests broader implications for multi-modal robot learning. Just as visual features benefit from camera calibration and geometric structure, tactile signals require proper spatial grounding to be maximally useful for control. We believe this principle extends beyond manipulation---any sensing modality that provides local measurements could benefit from explicit anchoring to relevant coordinate frames. As tactile sensors become more prevalent in robotics, frameworks like SaTA that preserve and exploit their geometric richness will be essential for pushing the boundaries of what robots can perceive and manipulate.

\section*{ACKNOWLEDGMENT}

This work was supported by the Shanghai Qi Zhi Institute and the Tsinghua University Dushi Program.

\end{document}